\documentclass{article}
\usepackage{hyperref}
\usepackage{url}
\usepackage{graphicx}
\usepackage{amsmath, amssymb, amsthm, epsfig, mathtools}

\newcommand{\beginsupplement}{%
        \setcounter{table}{0}
        \renewcommand{\thetable}{S\arabic{table}}%
        \setcounter{figure}{0}
        \renewcommand{\thefigure}{S\arabic{figure}}%
        \setcounter{section}{0}
        \renewcommand{\thesection}{\arabic{section}}%
     }
\usepackage{enumitem}
\setlist[itemize]{leftmargin=-5ex}
\setlist[itemize]{itemsep=-1ex}

\usepackage{wrapfig}

\title{A Useful Motif for Flexible Task Learning in an Embodied Two-Dimensional Visual Environment}

\PassOptionsToPackage{numbers,square}{natbib}

\usepackage[final]{nips_2017}

\usepackage[utf8]{inputenc} % allow utf-8 input
\usepackage[T1]{fontenc}    % use 8-bit T1 fonts
\usepackage{hyperref}       % hyperlinks
\usepackage{url}            % simple URL typesetting
\usepackage{booktabs}       % professional-quality tables
\usepackage{amsfonts}       % blackboard math symbols
\usepackage{nicefrac}       % compact symbols for 1/2, etc.
\usepackage{microtype}      % microtypography

\author{
Kevin T. Feigelis\\
Department of Physics\\
Stanford University\\
Stanford, CA 94305 \\
\texttt{feigelis@stanford.edu} \\
\And
Daniel L. K. Yamins \\
Departments of Psychology and Computer Science \\
Stanford Neurosciences Institute \\
Stanford University \\
Stanford, CA 94305 \\
\texttt{yamins@stanford.edu} \\
}

\begin{document}

\maketitle

\begin{abstract}
Animals (especially humans) have an amazing ability to learn new tasks quickly, and switch between them flexibly. 
How brains support this ability is largely unknown, both neuroscientifically and algorithmically.  
One reasonable supposition is that modules drawing on an underlying general-purpose sensory representation are dynamically allocated on a per-task basis.
Recent results from neuroscience and artificial intelligence suggest the role of the general purpose visual representation may be played by a deep convolutional neural network, and give some clues how task modules based on such a representation might be discovered and constructed. 
In this work, we investigate module architectures in an embodied two-dimensional touchscreen environment, in which an agent's learning must occur via interactions with an environment that emits images and rewards, and accepts touches as input. 
This environment is designed to capture the physical structure of the task environments that are commonly deployed in visual neuroscience and psychophysics.
We show that in this context, very simple changes in the nonlinear activations used by such a module can significantly influence how fast it is at learning visual tasks and how suitable it is for switching to new tasks. 
\end{abstract}

%Papers may be only up to eight pages long, including figures.
%An additional ninth page containing only cited references is allowed.

\section{Introduction} %~750 words
In the course of everyday functioning, animals (including humans) are constantly faced with  real-world environments in which they are required to shift unpredictably between multiple, sometimes unfamiliar, tasks~\cite{botvinick2014computational}.
They are nonetheless able to flexibly adapt existing decision schemas or build new ones in response to these challenges~\cite{arbib1992schema}.
How brains support such flexible learning and task switching is largely unknown, both neuroscientifically and algorithmically~\cite{wagner1998building}.   

One reasonable supposition is that this problem is solved in a modular fashion, in which simple modules~\cite{DBLP:journals/corr/AndreasRDK15}  specialized for individual tasks are dynamically allocated on top of a largely-fixed general-purpose underlying sensory representation~\cite{2017arXiv170405526H}. 
The general-purpose representation is likely to be large, complex, and learned comparatively slowly with significant amounts of training data.
In contrast, the task modules reading out and deploying information from the base representation should be lightweight and easy to learn.   
In the case of visually-driven tasks, results from neuroscience and computer vision suggest the role of the general purpose visual representation may be  played by the ventral visual stream, modeled as a deep convolutional neural network \cite{DiCarlo_2012, yamins:pnas2014}.
A wide variety of relevant visual tasks can be read-out with simple, often linear, decoders, based on features at combinations of levels in such networks~\cite{razavian2014cnn}.
However, how the putative ventral-stream representation might be deployed in an efficient dynamic fashion remains far from obvious. 

Work on lifelong (or continual) learning~\cite{DBLP:journals/corr/KirkpatrickPRVD16, DBLP:journals/corr/RusuRDSKKPH16}, reinforcement learning~\cite{DBLP:journals/corr/DuanSCBSA16,DBLP:journals/corr/JaderbergMCSLSK16,DBLP:journals/corr/WangKTSLMBKB16}, neural modules~\cite{DBLP:journals/corr/AndreasRDK15, 2017arXiv170405526H}, and decision making~\cite{Brunton95,doll2015model} have addressed many aspects of these questions, including how to learn new tasks without destroying the ability to solve older tasks, how to parse a novel task into more familiar subtasks, and how to determine when a task is new in the first place.   
In this work, we investigate a somewhat different question --- namely, how the local architectural motifs deployed in a module can influence how efficient a system is at learning and switching between tasks. 
We find that some simple motifs (e.g. low-order polynomial nonlinearities and sign-symmetric concatenations) significantly outperform more standard neural network nonlinearities (e.g. ReLus), needing fewer training examples and fewer neurons to achieve high levels of performance. 

Our work is situated in an \emph{embodied system}~\cite{anderson2003embodied} that is designed to capture the physical structure of the task environments that are commonly deployed in visual neuroscience and psychophysics.
Specifically, we model a two-dimensional touchscreen, in which an agent's learning must occur via interactions with an environment that emits images and rewards, and accepts touches as input. 
It has been shown that mice, rats, macaques and (of course) humans can operate touchscreens to learn operant behaviors~\cite{horner2013touchscreen,rajalingham_monkeybehavior_2015}.  
Building on work that has had success mapping neural networks to real neural data in the context of static visual representations~\cite{yamins2016using,cadieu2014deep,mcintosh2016deep}, our goal is to produce models of dynamic task learning that are sufficiently concrete that they can help build bridges between general theories and architectures from artificial intelligence, and specific experimental results in empirically measurable animal behaviors and neural responses.

\section{The Touchstream Environment} \label{Sec: touchstream}
\begin{figure}
\centering
\includegraphics [width=1\linewidth]{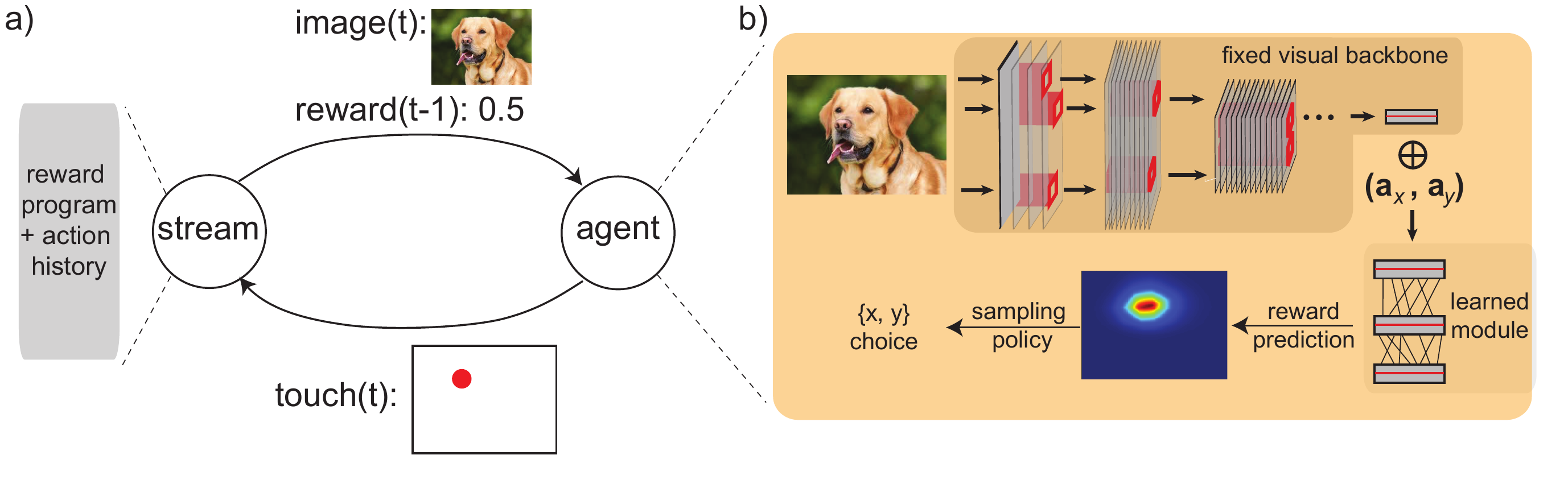}
\vspace{-5mm}
\caption{\textbf{The Touchstream environment.}  \textbf{a.} The Touchstream environment is a simple GUI-like testbed for posing visual tasks to a continual reinforcement learning agent. It consists of two interacting components, a screen server that emits images and rewards, and an agent that accepts images and rewards and emits touch actions, e.g. positions in a two-dimensional grid of the same shape as the input image.  \textbf{b.} The agent is a neural network with a fixed visual backbone and modules drawing on this backbone that learn to interact with the Touchstream so as to maximize long-term reward that is received.  On each timestep, the agent produces an estimate of rewards in the next one or several timesteps, conditional on the action it chooses at that timestep and in the recent past, in the form of a reward heatmap.  Using a simple exploration policy, the agent chooses its next action based on this heatmap.  In this work, our goal is evaluate what types of module architectural motifs enable the agent to efficiently learn new Touchstream tasks.
\label{fig:touchstream}}
\vspace{-5mm}
\end{figure}

The Touchstream environment consists of two components, a screen server and an agent, interacting over an extended temporal sequence (Fig. \ref{fig:touchstream}a). 
At each timestep $t$, the screen server emits an image $i_t$ and a non-negative reward $r_t$.  Conversely, the agent accepts images and rewards as input and on the next timestep emits an action $a_t$ in response.  
The action space available to the agent consists of a pixel grid of the same shape as the input image.  
The screen server runs a program computing $i_{t}$ and $r_{t}$ as a function of the history of agent actions $\{a_0, \ldots, a_{t-1}\}$, images $\{r_0, \ldots, i_{t-1}\}$ and  rewards $\{r_0, \ldots, r_{t-1}\}$. 
The agent is a neural network (Fig. \ref{fig:touchstream}b), composed of a visual backend with fixed weights, together with a recurrent module whose parameters are learned by interaction with the Touchstream (see section \ref{sec:architectures} for more details on model architectures) .

This setup is meant to mimic a touch-based GUI-like environment such as those used in visual psychology or neuroscience experiments involving both humans and non-human animals~\cite{horner2013touchscreen,rajalingham_monkeybehavior_2015}.   
The screen server is a stand-in for the experimentalist, showing visual stimuli and using rewards to elicit behavior from the subject in response to the stimuli. 
The agent is analogous to the experimental subject, a participant who cannot receive verbal instructions but is assumed to want to maximize the aggregate reward it receives from the environment.   
In principle, the screen server program can be anything, encoding a wide variety of two-dimensional visual tasks or dynamically-switching sequences of tasks.
In this work, we evaluate a small set of tasks analogous to those used in simple human or monkey experiments, including Stimulus-Response and Match-to-Sample categorization tasks, and object localization tasks.  

\begin{figure}
\centering
\includegraphics [width=1\linewidth]{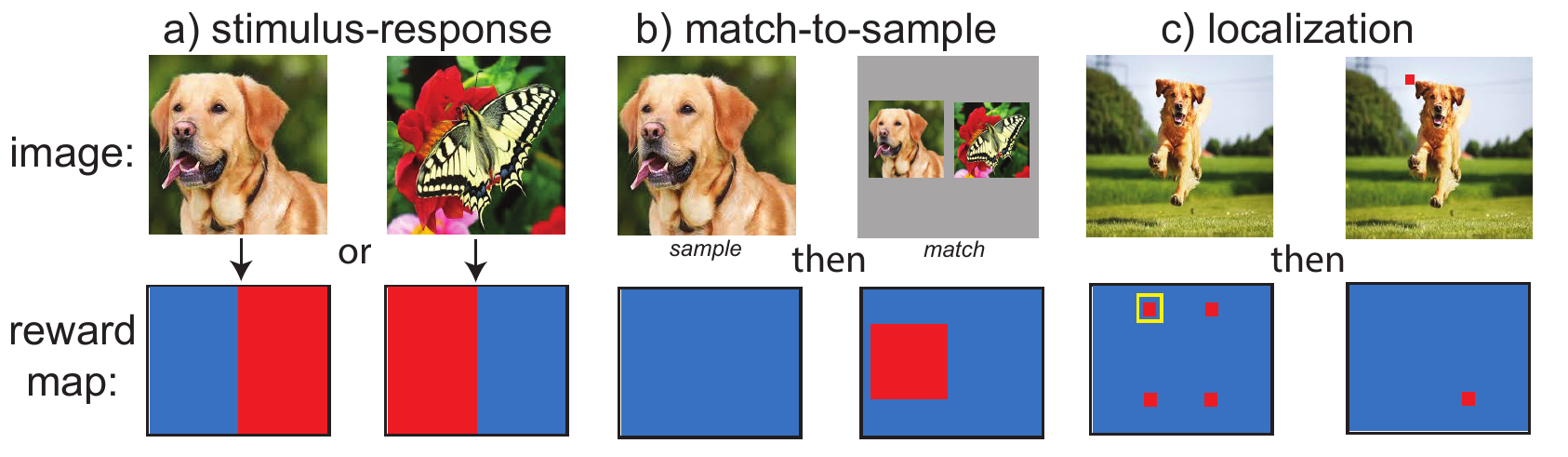}
\vspace{-5mm}
\caption{\textbf{Tasks.}  Illustration of how the taskstream environment captures a variety of tasks commonly used in visual psychophysics experiments.  The top row shows the image output of the screen server, while the bottom illustrates the reward map for the task, with red indicating high reward and blue indicating low reward.  \textbf{a.} A binary Stimulus-Response task in which the reward function is defined by associating the left half of the screen with the first class, and the right hand side of the screen with the second class.
The agent recieves an image frame containing one of the two classes, and must respond by touching the image on the correct side to receive a reward.
This type of SR task can be generalized to more classes by, e.g., further sectioning of the screen. 
\textbf{b.} The Match-To-Sample task displays a \emph{sample} stimulus screen of a single class instance to the agent, which may be touched anywhere to proceed to the following \emph{match} screen.
This next screen displays several smaller images, one of which depicts the same class as the sample screen, while the rest are distractor classes.
The agent is rewarded for touching within the correct class box.
\textbf{c.} Localization is a single image containing an instance of one unique class. The agent is rewarded by first touching two opposing corners of the object, which defines a bounding box.   The correct touch on the second screen depends on the action chosen on the first screen, which is indicated with a small yellow box (this box is purely illustrative and not actually part of the reward map).  \label{fig:taskstreams}}
\vspace{-6mm}
\end{figure}

\textbf{Stimulus-Response Tasks:}  The Stimulus-Response (SR) paradigm is a simple way to physically embody discrete categorization tasks that are commonly used in the animal (and human) neuroscience literature~\cite{Gaffan1988149}. 
In the Touchstream environment, the $N$-way SR task assumes that input images are divided into $N$ classes, and that the screen is correspondingly divided into $N$ regions $b_1, \ldots, b_N$.  
A reward of 1 is returned if the agent's touch is inside the region associated with the class identity of shown image (e.g. $a_t \in b_{class(i_t)})$, and 0 otherwise.
For example, in a two-way SR discrimination task (Fig. \ref{fig:taskstreams}a), the agent might be rewarded if it touches the left half of the screen after being shown of a dog, and the right half after being shown an image of a butterfly.
The SR task can be dialed in difficulty by increasing the number of image classes or the complexity of the class regions. 
Multiple classes can be introduced all at once, or curricularization could be achieved by (e.g.) slowly interleaving new classes into the corpus while the reward boundaries are morphed accordingly.
The present study examines the following three SR variants: two-way binary classification with left/right decision boundaries, four-way double-binary classification in which two pairs of classes define two independent left/right decision boundaries, and four-way classification in which the reward surface splits the Touchstream GUI into quarters.
Image categories used are drawn from the Image-Net 2012 ILSVR classification challenge dataset \cite{imagenet_cvpr09}.  Each class has 1300 unique training instances, and 50 unique validation instances. 

\textbf{Match-To-Sample Tasks:} Another common approach to assessing visual categorization abilities is the Match-To-Sample (MTS) paradigm~\cite{Murray6568}.
An MTS trial is a pair of image frames (Fig. \ref{fig:taskstreams}b).  A trial begins by presenting the agent with a unique `sample' screen depicting an image of a single class instance.
The agent is allowed to touch anywhere within the GUI to advance to the next frame and receives no reward for this frame.
Presented next is the 'match' screen, which displays to the agent multiple small images on a blank background, each containing a fixed template image of one of $N$ classes, one of which is of the class shown on sample screen.
After this frame, the server returns reward of 1 if the agent touches somewhere inside the rectangular region occupied by the image of the class shown in the first image, and 0 otherwise.
The MTS paradigm incorporates the need for working memory and more localized structure.
Along with standard binary discrimination, we consider variants of the MTS task which make the match screen less stereotypical, either with more distractor classes, with random vertical translations of the match images, with random interchanging of the lateral placement of the two classes, or all perturbations simultaneously (see supplementary materials for more information about specific tasks).

\textbf{Localization Tasks:}
In both the SR and MTS tasks, the reward value of a given action is independent of actions taken on previous steps.
To go beyond this situation, we also explore a two-step localization task (Fig. \ref{fig:taskstreams}c), using synthetic images containing a single main salient object placed on a complex background (similar to images used in \cite{yamins2016using,yamins:pnas2014}). 
Each trial of this task has two steps, and the agent is rewarded after the second step by an amount proportionate to the Intersection over Union (IoU) value of box implied by the agent's touches, relative to the ground truth bounding box.
No reward is dispensed on the first touch of the sequence.

\textbf{Bounded Memory and Horizon:} The tasks described above require bounded memory and have a fixed horizon.  
That is, a perfect solution always exists within $k_{f}$ timesteps from any point, and requires only knowledge of the previous $k_{b}$ past steps, for $k_b \leq 1$ and $k_f \leq 2$.  
(Specifically, $k_b=0, k_f=1$ for SR, $k_b=k_f=1$ for MTS, and $k_b=1, k_f=2$ for localization).
For more complex visual tasks such as object segmentation, these numbers will be larger (and potentially unbounded) or vary in a complex way between trials, but for the present work we avoid those complications. 

\section{Module Architectures} \label{sec:architectures}
In what follows, we will denote the two-dimensional pixel grid action space as $\mathcal{A}$.
We define a \emph{module} $M$  to be a neural network whose inputs are a history over $k_b$ timesteps of (i) the agent's actions, and (ii) the activations of a fixed visual backbone model; and which outputs reward prediction maps across action space for the next $k_f$ timesteps.  That is:
$$M_{P}: [\mathbf{C}_t, \mathbf{h}_{t-1}] \longmapsto (m_1, m_2, \ldots, m_{k_f})$$
where $\mathbf{C}_t$ is the history $(C_{t-k_b}, \ldots, C_t)$ of outputs of the backbone visual model, $\mathbf{h}_{t-1}$ is the history $(a_{t-k_b} \ldots, a_{t-1})$ of previously chosen actions,
and each $m_i \in \mathbf{map}(\mathcal{A}, [0, 1])$  --- that is, a map from action space to reward space. $P$ are the learnable parameters of the module network. 
In this work, we chose the visual backbone model be the output of the fc6 layer of the VGG-16 network pretrained on ImageNet~\cite{simonyan2014very}. 
The parameters $P$ of the modules are learned by stochastic gradient descent on reward prediction error, with map $m_j$ compared to the true reward $j$ timesteps after the action was taken.
\begin{wrapfigure}{L}{0.6\textwidth}
\centering
\includegraphics[width=0.6\textwidth]{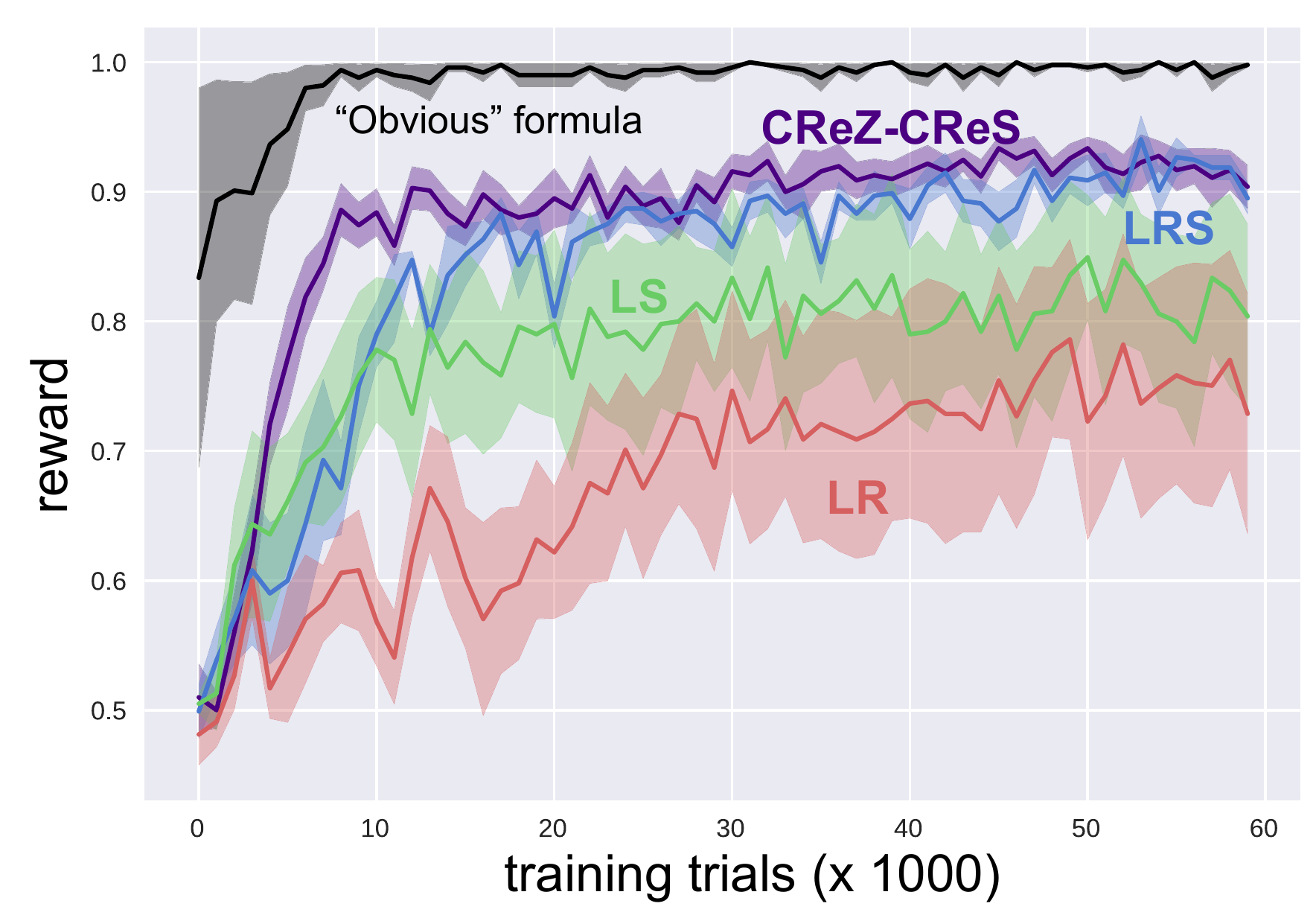}
\caption{\footnotesize{The ``obvious'' solution to a two-way SR task, using the formula in eq. \ref{eq: SR perfect}, leads to the black line learning curve shown above.  
Results for the more generic early-bottleneck modules motifs are shown for comparison.}\label{fig: perfect_SR}}
\vspace{-5mm}
\end{wrapfigure}

Having produced a reward prediction map, the agent chooses its action for the next timestep by  normalizing the predictions across the action space on each of the $k_f$ timesteps into $k_f$ separate probability distributions, and sampling from the distribution in the frame that has maximum variance (see supplementary info for details). We chose this policy because it lead to better final results on our very large action space in comparison to a large variety of more standard reinforcement learning policy alternatives (e.g. $\epsilon$-greedy), although the rank order of the comparison results reported below appeared to be robust to choice of action policy.

Although the minimimal required values of $k_b$ and $k_f$ are different from the various tasks in this work, in all the investigations below, we take $k_b = 1$ and $k_f = 2$ --- the maximum values of these across tasks --- and require architectures to learn on their own when it is safe to ignore information from the past and when it is irrelevant to predict past a certain point in the future. 

\textbf{``Obvious Solutions''} 
The main question we seek to address here is: what architectural motif should the module $M_P$ be? 
The key considerations for such a module are that it should be (i) easy to learn parameters $P$, requiring comparatively few training examples, (ii) easy to \emph{learn from}, meaning that a new task related to an old one should be able to more quickly build its module by reusing components of old ones, and (iii) general, e.g. having the above two properties for all or most tasks of interest. 
How should such a structure be found? 

As an intuition-building example, it's useful to note that many of the tasks we address here have ``obvious solutions'' --- that is, easily written-down analytic expressions that correctly solve the task. 
Consider the case of a 2-way SR task like the one in Fig. \ref{fig:touchstream}a. 
Given verbal instructions describing this task, a human might solve this task by allocating a module to compute the following function:
\begin{equation}\label{eq: SR perfect}
M[C](a_x) = \mathbf{ReLu}(WC)\cdot\mathbf{ReLu}(-a_x) + \mathbf{ReLu}(-WC)\cdot\mathbf{ReLu}(a_x)
\end{equation}
where $a_x$ is the $x$-component of the action $a \in \mathcal{A}$, and $W$ is a length-$|C|$ vector expressing the dog/butterfuly class boundary (bias term omitted for clarity).
Since $k_b=0$ for this task, the formula doesn't depend on previous action history $\mathbf{h}$ at all. 
Given the ability to recall $W$ from long-term memory, this formula will allow the human to solve the task nearly instantly. 
In fact, even if the weight matrix is not known, imposing the structure in eq. \ref{eq: SR perfect} allows $W$ to be learned quickly, as illustrated by the learning curve shown as a black line in Fig. \ref{fig: perfect_SR}.   
By narrowing down the decision surface considerably with a ``right'' decision structure, the learnable parameters of the module must simply construct the category boundary, which, for a good visual feature representation (such as VGG-16), is comparatively trivial and converges to a nearly-perfect solution extremely quickly.

\begin{figure}
\centering
\includegraphics [width=.8\linewidth]{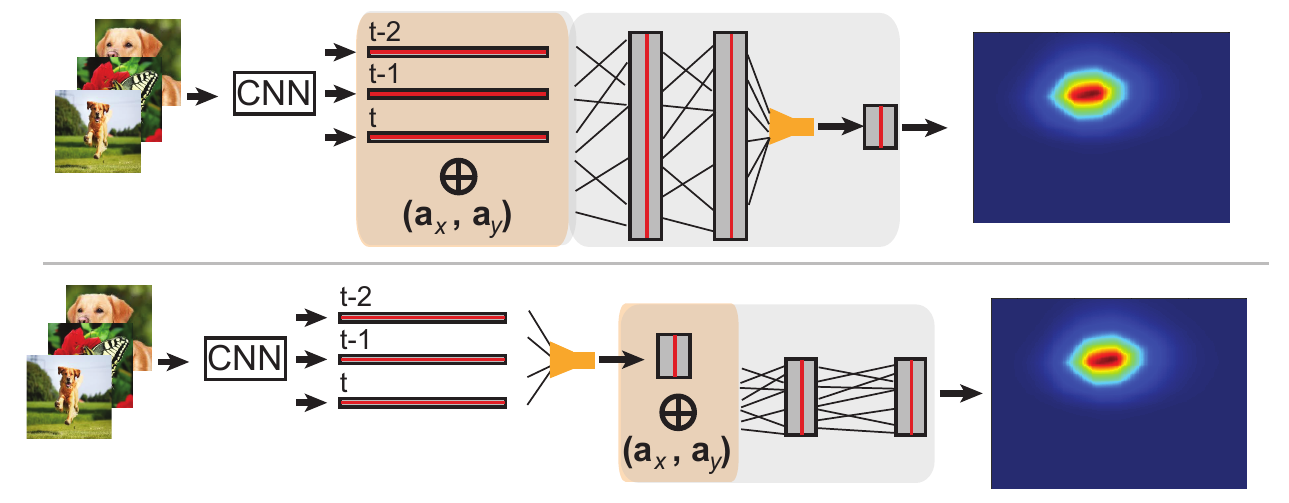}
\vspace{-3mm}
\caption{\textbf{a.} Late bottleneck (LBX) module which concatenates actions directly with visual features. \textbf{b.}  Early bottleneck module, which reduces the number of visual dimensions before action concatenation. \label{fig: Architectures}}
\vspace{-5mm}
\end{figure}

The agents modeled in the present work do not have have the language capabilities to receive verbal instruction that direct the allocation of ``obvious'' task-specific layout structures so flexibly.  
Nonetheless, it is possible to glean a few basic ideas from this example that are  generally useful (Fig. \ref{fig: Architectures}).  
First, the module has an \emph{early bottleneck}; that is, the high-dimensional general feature representation is reduced to a small number of dimensions (in this case, from 4096 to 1) before being combined with the action parameter $a$.   
Second, the module is \emph{multiplicative}, in that interaction between action features and visual features involves a multiplication operation. 
Thirdly, the module has two sign-symmetric terms, accounting for the flip symmetry in the action-reward relationship across the vertical axis. 
It turns out that early bottlenecks, multiplication and sign-symmetry can be built into a very simple general design that can be learned entirely from scratch, leading to efficient learning not just for the binary SR task but for a variety of visual decision problems.

\textbf{Generic Motifs}  We define several possible generic motifs which implement some or all of these three basic ideas.  Specifically:

\begin{itemize}[leftmargin=*,itemsep=0ex,topsep=1ex]
\item \textbf{The LRS Module:}  This is a standard MLP using a nonlinearity that concatenates a multiplicative square term to the standard ReLu nonlinearity, e.g. 
$$x \mapsto \mathbf{ReLu}(x) \oplus x^2 \coloneqq \mathbf{RS}(x)$$ where $\oplus$ denotes vector concatenation.
Thus, after one layer, the features are of the form $ReLu(W \cdot x + b) \oplus (W \cdot x + b)^2$ for some (learnable) tensors $W, b$.  
After two layers of the $LRS$ motif, cross-products emerge between the elements of the original input $x$. 
We define the $(n_0, n_1, \ldots, n_k)$-LRS module as the network which bottlenecks its input to dimension $n_0$, combines with actions, and then performs $k$ LRS layers of size $n_1, \ldots, n_k$, respectively.  That is, 
$$z = \mathbf{ReLu}(W_0 \cdot C + b_0)$$
$$\quad l_1 = \mathbf{RS}(W_1 (z \oplus a) + b_1)$$
$$\quad l_i = \mathbf{RS}(W_i l_{i-1} + b_i) \quad\text{for}\quad i > 1$$
where $W_i$ is of shape $(n_{i-1}, n_i)$, and $b$ is of shape $n_i$.
The core motivation for the LRS motif is that the ``obvious'' architecture for the SR task, described above, is special case of a $(1, 2, 2)$-LRS module.  

\item \textbf{The LX Modules:}  For comparison and control, we define ``LX'' modules to be same as the LRS module, but with the \textbf{RS} activation function replaced by various other, mostly more standard, nonlinear activation functions (the ``X''s).  For example, we test the squaring nonlinearity by itself without the \textbf{ReLu} concatenated, which we denoted the $LS$ module.  We also test the \textbf{ReLu} nonlinearity alone, denoted as the $LR$ module, as well as a $\tanh$ nonlinearity (the $LT$ module), and a sigmoid nonlinearity (the $LSig$ module).
Additionally, we also test more recently explored activations such $elu$ (the $LE$ module)~\cite{DBLP:journals/corr/ClevertUH15}, and the Concatenated \textbf{ReLu} (the $LCRe$ module)~\cite{DBLP:journals/corr/ShangSAL16} are tested.

\item \textbf{The LBX Modules:} ``Late-Bottleneck'' modules using the same nonlinearities ``X'' as above, except actions are combined with visual features before any bottlenecking has been done.
This is just a standard MLP in which actions are concatenated onto the visual features at the beginning.
If the visual backbone model output is of size $N$, the input to such a module will be of size $N+2$ per timestep, since action space $\mathcal{A}$ is 2D.

\item \textbf{The CReZ-LRS Module:} An LRS module where the \textbf{ReLu} nonlinearity in the bottleneck layer (denoted $z$ above) is replaced by the \textbf{CReLu} nonlinearity defined by $x \mapsto \mathbf{ReLu}(x) \oplus \mathbf{ReLu}(-x)$.
Motivating this addition to the LRS motif, is that the symmetric nature of the ``obvious'' SR architecture is captured explicitly through this sign-symmetric transformation on visual features.

\item \textbf{The CReZ-CReS Module:} This uses the same \textbf{CReLu} bottleneck nonlinearity as above, except the $RS$ activations in the following layers are replaced by a nonlinearity which concatenates the squares of each of the CReLu components, e.g.
$$x \mapsto \mathbf{ReLu}(x) \oplus \mathbf{ReLu}(-x) \oplus \mathbf{ReLu}^2(x) \oplus \mathbf{ReLu}^2(-x) \coloneqq \mathbf{CReS}(x)$$
The LRS motif can be seen as a subset of this module, using only the first component of the \textbf{CReS} nonlinearity, and whichever square term is nonzero following rectification.

\end{itemize}

From the point of the Universal Approximation Theorem (UAT)~\cite{cybenko1989approximation}, these motifs have equivalent expressivity, in that one could approximate (e.g.) the square non-linearity with a sufficiently large LBX module. 
However, the UAT makes no guarantees about learning efficiency or network size. 
Crucially for the results of paper, the types of nonlinearities needed to capture task demands in the Touchstream environment make some of these motifs significantly more efficient than others. 

\section{Experiments}
We compared the LRS, LX, LBX, CReZ-LRS, and CReZ-CReS modules on twelve variants of the SR, MTS, and localization tasks.
Weights were initialized using the Xavier algorithm~\cite{glorot2010understanding}, and were learned using the ADAM optimizer~\cite{DBLP:journals/corr/KingmaB14} with parameters $\beta_1=0.9$ and $\beta_2=0.999$ and $\epsilon=1e-8$.   
Learning rates were optimized on a per-task per-architecture basis in a cross-validated fashion. 
For each architecture and task, we ran optimizations from five different weight initializations to obtain mean and and standard error due to initial condition variability. 
For the LBX modules, we measured the performance of modules of three different sizes (small, medium and large).  
Throughout, the small LBX version is equivalent in size to the small early bottleneck modules, whereas the medium and large versions are much larger. 

\begin{figure}
\centering
\includegraphics [width=1\linewidth]{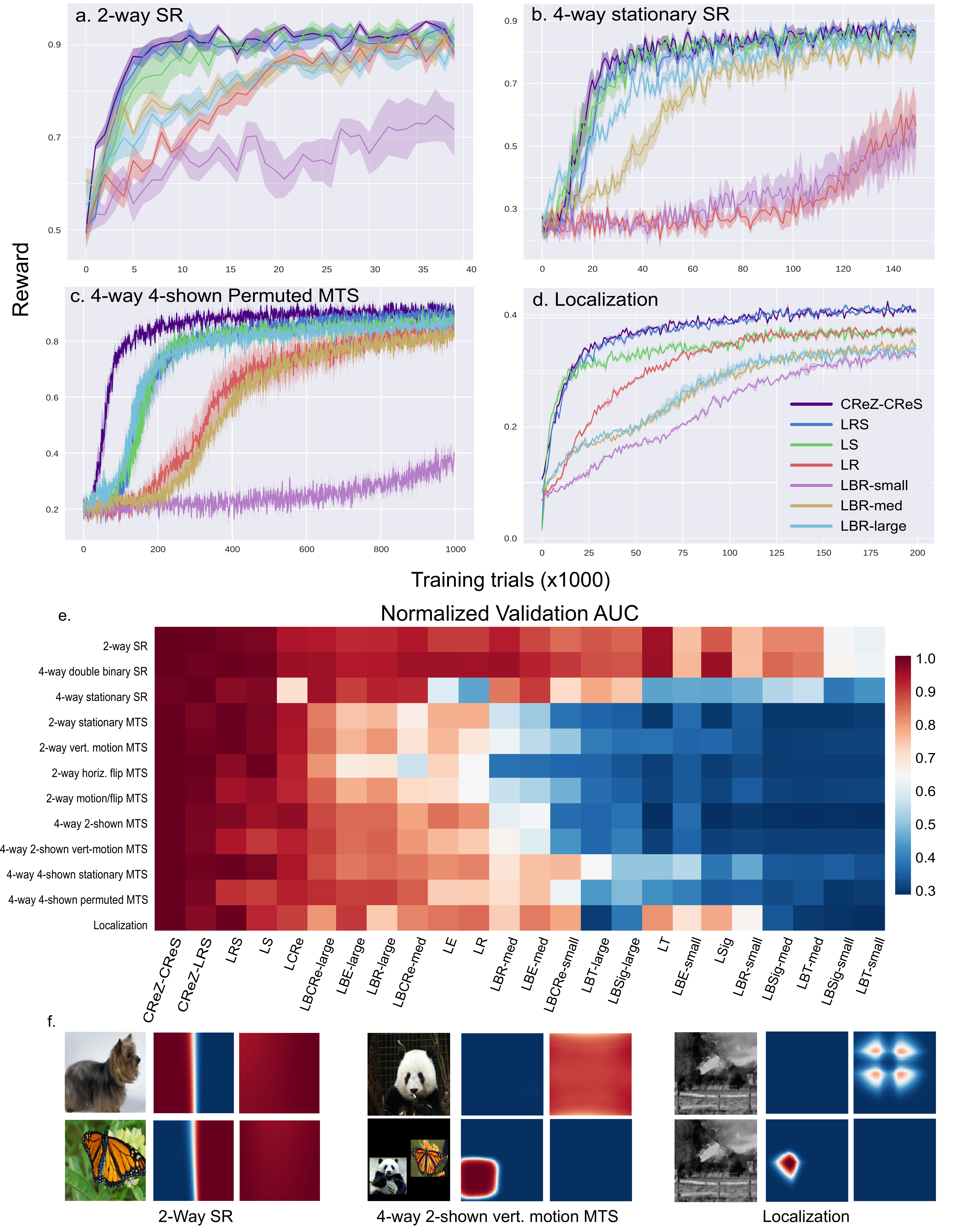}
\vspace{-3mm}
\caption{\textbf{Learning curves.} Reward obtained over the course of training for modules on  \textbf{a.} two-way Stimulus-Response, \textbf{b.} four-way SR, \textbf{c.} four-way MTS with four randomly ordered match options, and \textbf{d.} Localization. Solid lines represent mean and shaded area represents standard errors, taken over training runs with different weight initializations.  For clarity, only a subset of the tested nonlinearities are displayed. \textbf{e.} Area Under the learning Curve (AUC) for all twelve task variants of each task we implemented, normalized by the value obtained for the highest performing model. Modules are shown ordered by mean performance across tasks.  \textbf{f.} Heatmaps for current and next time steps learned by LRS module for two-way Stimulus-Response (left), four-way two-shown Match-To-Sample with random vertical motion (center), and Localization (right). 
\label{fig:Results_1}}
\vspace{-5mm}
\end{figure}

Our main results are that:
\begin{itemize}[leftmargin=*,itemsep=0ex,topsep=1ex]
	\item Small early-bottleneck modules using the squaring operation in their nonlinearity (CReZ-CReS, CReZ-LRS, LRS, and LS) learn the tasks substantially more quickly than the other architectures at comparable or larger sizes, and often attain a higher final performance levels. 
	\item The small early-bottleneck module with the sign-symmetric early nonlinearity (LCre) is less efficient than modules with the square nonlinearity, but is substantially better than architectures with neither the early bottleneck nor the square nonlinearity. 
	\item The CReZ-CReS module, which combines the early bottleneck, the squaring nonlinearity, and sign-symmetric concatenation, is the most efficient on all but one task (Fig. \ref{fig:Results_1}), while the CreZ-LRS module, which is generally the second most efficient, wins in one task by a very small margin. 
\end{itemize}
\noindent In other words, the main features that lead the ``obvious'' architecture to do well in the binary SR task are both individually helpful, and combine usefully, across a variety of visual tasks. 

\textbf{SR Experiment Details:} 
For this task the CReZ-CReS, CReZ-LRS, LRS, LX, and small LBX architectures had 8 units per layer, while the medium and large LBX had 128 and 512 units, respectively. 
We find that on Stimulus-Response Tasks with left versus right decision boundaries (Fig. \ref{fig:Results_1}a), all models aside from the smallest LBX eventually achieve reasonably high maximum reward.
Typically, we find that the (small-sized) non-squaring LX modules have performance better than the medium-sized but worse than the large-sized LBX modules, except in the case of the four-way quadrant variant of the SR task, where all non-squaring LX fail to converge to a solution within the alotted timeframe (Fig. \ref{fig:Results_1}b).
In contrast, the small-sized CReZ-CReS, CReZ-LRS, LRS, and LS modules however learn this task with few training examples, resulting in reward prediction maps such as those in Fig. \ref{fig:Results_1}f (left).

\textbf{MTS Experiment Details:} 
For this task the CReZ-CReS, CReZ-LRS, LRS, LX, and small LBX architectures had 32 units per layer, while the medium and large LBX had 128 and 512 units, respectively.
Fig. \ref{fig:Results_1}c offers a glimpse at characteristic learning curves observed for more challenging tasks (e.g. a four-way MTS variant).
CreZ-CReS, LRS ,and LS are seen to achieve peak performance within the initial stages of learning, whereas all other modules follow a sigmoid-like trajectory.
Non-sterotyped match screens are observed to present difficulties to the small and medium LBX modules, in contrast to the squaring modules which solve these as efficiently as stationary versions while maintaining a high degree of precision (Fig. \ref{fig:Results_1}f, middle).
We note from both MTS and SR heatmaps that at the beginning of a trial, the LRS is confident in its ability to receive a reward at time $k_f=2$ with zero knowledge of what the next frame may be.

\textbf{Localization Experiment Details:} 
In this task the CReZ-CReS, CReZ-LRS, LRS, LX, and small LBX architectures had 128 units per layer, while the medium and large LBX had 512 and 1024 units, respectively.
IoU curves in Fig. \ref{fig:Results_1}d show little difference between any of the LBR models despite size, and that the only models to consistently achieve an IoU above 0.4 are the CReZ-CReS, LRS, and CReZ-LRS (unshown).
To give context for the IoU values, we note all of the early-bottlenecked modules are able to equal or outperform a baseline SVR trained using supervised methods to directly regress bounding boxes using the same VGG features (0.369 IoU).
Inspecting the prediction heatmaps for the LRS module in this task (Fig. \ref{fig:Results_1}f, right) shows that the reward uncertainty frontier is well-localized.

\textbf{Basic Task-Switching:} Next, we tested these modules under basic task-switching conditions to determine their suitability for redeployment.
We chose to contrast the performance of the LRS and largest LBR modules which were both initially trained on the two-way SR task.
These are then repurposed to solve the same task but with class boundaries reversed, the same task but with two entirely new categories, and the four-way double binary version.
We repeat this experiment using the two-way MTS task as well.
After switching, we hold all $k_f$ output map weights fixed for both models.
The LRS further has its $n_1$ and $n_2$ layer weights fixed, whereas the LBR is tested both with and without holding its second fully-connected layer constant.

\begin{figure}
\centering
\includegraphics [width=1\linewidth]{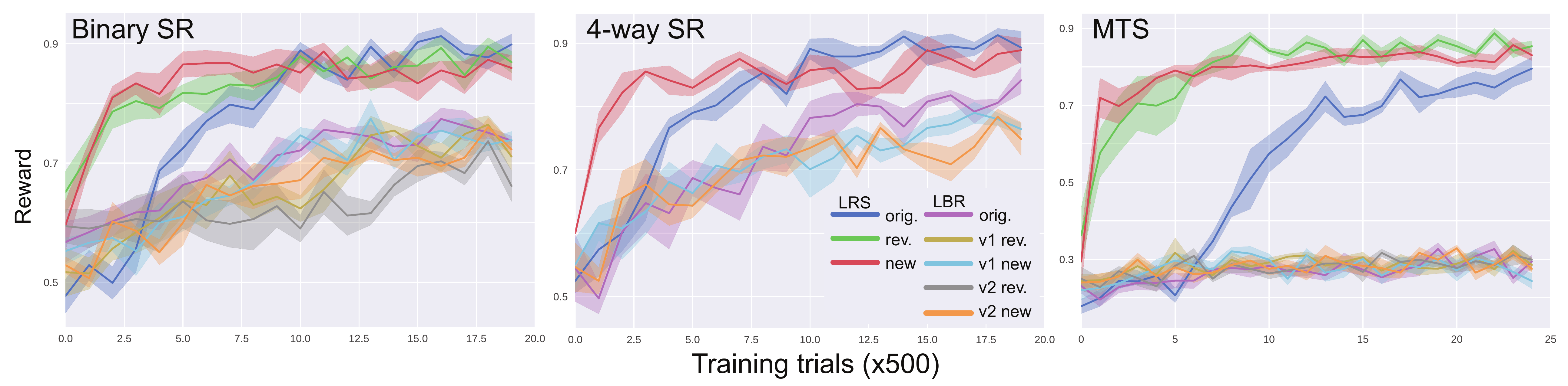}
\vspace{-6mm}
\caption{\textbf{Basic Task Switching} in SR and MTS tasks when the class boundaries are either reversed or two new classes introduced.  \textbf{a. Binary SR task.}  For the binary SR task, blue line (``orig.'') is the LRS module's original training curve, green line  (``rev.'') indicates the training curve for the LRS module with reversed-class boundary after task switching, while red line (``new'') indicates the training curves for the LRS module with a new category boundary.  Also shown are reversed and new boundary task switching results for late-bottleneck Relu (LBR) module, with either both FC layers in the module relearned (``v1'') or just the first layer relearned with the second layer is held fixed (``v2'').  \textbf{b.} Analogous results for a 4-way SR task; note that no ``reversed'' condition exists for this task. \textbf{c.} Analogous results for the binary MTS task. 
\label{fig:Results_3}}
\vspace{-5mm}
\end{figure}

In all cases the LRS is immediately able to adapt to the new class boundaries and learn the new tasks quicker than one trained from scratch (Fig. \ref{fig:Results_3}).
Consistent with results in other contexts, the LBR module however is unable to easily use its prior experience in learning the new task, while holding its second FC layer constant actually hinders its ability to learn the new tasks.
These patterns are maintained for the more complex switch to double binary task (Fig. \ref{fig:Results_3}b) and the Match-To-Sample task (Fig. \ref{fig:Results_3}c), with the large LBR model even failing to converge within the short timeframe needed to study switching behavior.
In conclusion, we find that the mapping functions developed by the LRS on a base task may be rapidly transfered to new LRS modules learning tasks with similar reward functions.

As yet, we have not yet tested the CreZ-CreS or CreZ-LRS modules in the task switching context, nor have we performed any experiments with task-switching in the context of the localization task.   We plan to do these experiments in the near future.

\section{Conclusion and Future Directions}  %~250 words
In this work, we introduced a two-dimensional embodied framework that mimics the physical setup of an experimental touchscreen environment, with the eventual goal of allowing direct comparison between learning characteristics of embodied models, and real animals in neuroscience experiments, which are necessarily embodied.
We showed that simple module structures building on a fixed visual backbone can learn a selection of tasks posed during typical visual neuroscience experiments, using a basic reinforcement learning scheme.
Moreover, we found that the choice of nonlinearity in the module can have significant effects on the ability of the module to learn quickly and switch tasks efficiently. 
This is fundamentally because, in our task space, many ``natural'' interactions between action space and visual features are multiplicative.
To allow module structures to remain small (and thus quickly learned and repurposed for new tasks), it is useful for this natural multiplicative interaction to be directly available in the module architecture. 

The fact that the squared and concatenated sign-symmetric nonlinearities (LS, LRS, CreZ-LRS and CreZ-CreS) are superior to other nonlinearities in our experiments is likely tied to the spatial structure of the two-dimensional embodied environment in which it is situated.  
Real neuroscience or psychophysics experiments are actually embodied, and our results suggest that the nature of the embodiment might be an important consideration when making detailed models of real experimental data.
However, the relevance of our results to other more general task-learning and switching situations is, by the same token, limited by the extent to which other decision and task switching processes are themselves embodied in spatial-like environments.  

It is very likely that other architectures are even better than CReZ-CReS.  
For instance, it is plausible that modules that generate convolutional weights that can be applied across images might lead to better localization results. 
In general, expanding to other module structures such as those as in \cite{DBLP:journals/corr/AndreasRDK15,2017arXiv170405526H,2017arXiv170503633J} would be of interest.
Finding a way to incorporate text inputs (that is, ``experimental instructions''), would also be of significant interest: the early gap between the blue and black learning curves Fig. \ref{fig: perfect_SR} illustrates how important the linguistic advantage remains (the late gap being due in part to the suboptimal action sampling policy used by generic architectures.)

While finding better module motifs is an important task, an equally crucial future step from a purely computational point of view will be to integrate insights from this work into more full-stack continual-learning procedure.  
Specifically, future work will need to address issues of: more sophisticated robust task-switching beyond the very simple tests we've done here; \emph{when} to declare a new module needed (presumably when a previously stable high-confidence module suddenly stops predicting the reward); how to \emph{grow} new modules in an efficient manner relative to existing modules~\cite{DBLP:journals/corr/RusuRDSKKPH16, DBLP:journals/corr/ChenGS15}; and how to consolidate modules when appropriate (e.g. when a series of tasks previously understand as separate can be solved by a single smaller structure). 

Though CReZ-CReS may be more effective than other motifs on the metrics we measure, that does not necessarily mean that is better description of what really happens in brains.
To determine if it is, a core goal core for our future work will need to involve: 
\begin{itemize}[leftmargin=*,itemsep=0ex,topsep=1ex]
 \item obtaining behavioral data from human and animals, and comparing it to the predictions made by the various motifs, seeing which (if any) best predicts learning rate and curve shape patterns in the data, using techniques like those deployed to match non-embodied model behavior to monkey and human behavior~\cite{rajalingham_monkeybehavior_2015}, and/or 
 \item obtaining neural data from animals and comparing response patterns to internal states of the model over learning time, extending techniques like those used in modeling the ventral visual stream~\cite{yamins2016using,yamins:pnas2014}.
\end{itemize}
\noindent It will be especially interesting to compare data between two species with obviously differential levels of task learning flexibility, e.g. mice vs rats (using techniques like  those in \cite{rogerson_sfn_2016}) or monkeys vs humay (using techniques like those in \cite{rajalingham_monkeybehavior_2015}), to see whether differences in local computational circuit architecture might explain intraspecies differences.

{\small
\bibliography{refs2.bib}}
\bibliographystyle{plain}

\section*{Supplementary Material}
\beginsupplement
\appendix
\section{Datasets} \label{Sec: Datasets}
Four unique object classes are taken from the Image-Net 2012 ILSVR classification challenge dataset to be used in Stimulus-Response and Match-To-Sample experiments: Boston Terrier, Monarch Butterfly, Race Car, and Panda Bear.
Two-way classification on these tasks uses Boston Terriers and Monarch Butterflies, except in switching experiments where the models are repurposed to classify Race Cars versus Panda Bears.

A synthetic dataset containing 100,000 unique training images and 50,000 unique validation images was generated for the Localization task.
This consisted of 59 unique classes of 3-D objects rendered on top of complex natural scene backgrounds.
Each image is generated using a one of these objects with randomly sampled size, spatial translation, and orientation with respect to a random background.

All images are preprocessed by taking a randomly cropped 224x224 segment, and subtracting the mean RGB pixel values for the 2012 ILSVR classification challenge dataset.

\section{MTS Methods} \label{Sec: MTS Methods}
Class template images for the match screen were held fixed at 100x100 pixels.
These contain a stereotypical face-centered and centrally-located example of the class in question.
For all variants of the MTS task, we keep a six pixel buffer between the edges of the screen and the match images, and a twelve pixel buffer between the adjascent edges of the match images themselves.
Variants without vertical motion have the match images vertically centered on the screen.

For two-way classification, we make the task more challenging in three ways.
We consider (i) translating each match image by a random number of pixels (but so that the image still is within the buffer), (ii) randomly flipping the horizontal location of the match images between trials, and (iii) both of these in tandem. 

Four-way classification has variants which either show two or four class options on the match screen.
If the match screen contains two classes, they are always randomized horizontally (otherwise the agent would always be rewarded for touching a single side).
In addition to this task, we also study a more challenging variant by introducing similar vertical motion as above.
On four-shown variants, we can increase complexity by randomly permuting the match screen orderings between trials. 

\section{Task Evaluation Metrics} \label{Sec: Task Evaluation Metrics}

Five runs per model were conducted on each task.
Every 1000 training trials, the model's performance on SR and MTS tasks was taken as the average reward obtained on the validation set.
Localization calculates its reward as the Intersection over Union (IoU) value between ground truth bounding box $B$ and predicted bounding box $\hat{B}$ as
$$IoU=\frac{Area(B \cap \hat{B})}{Area(B \cup \hat{B})}$$
No reward is given to the agent if the IoU value is below 0.5.
Rewards and IoUs are averaged across runs for each model, and Area under the Curve (AUC) values were calculated using a trapezoidal integration routine on the valitidation reward trajectories. 

\section{Modules} \label{Sec: Modules}
\subsection{Learnable Parameter Count}
Table \ref{Tab: Parameters} lists the total number of learable parameters for each model across all three base task paradigms. "X" Architechtures include squaring, ReLu, tanh, sigmoid, and elu actvations. CReLu has a different number of parameters due to concatenation.

\begin{table}
\centering
  \caption{Number of module parameters}
  \tiny{
  \hspace*{-0.5cm}
    \begin{tabular}{l *{13}{p{0.8cm}}}
    
        Base-task & CReZ-CReS & CReZ-LRS & LRS    & LX   & LCre & LBX-Small & LBX-Med  & LBX-Large & LBCre-Small & LBCre-Med  & LBCre-Large \\ \hline
        SR        & 66,108 & 65,916 & 65,852   & 65,756   & 65,852 & 65,684    & 1,066,244 & 4,461,572 & 65,780 & 10,83140 & 4,725,764 \\ 
        MTS       & 269,028 & 266,724 & 265,700  & 264,548  & 265,700 & 263,492   & 1,066,244 & 4,461,572  & 264,644 & 1,083,140 & 4,725,764\\ 
        LOC       & 1,149,828 & 1,116,036 & 1,102,222 & 1,084,046 & 1,099,652 & 1,067,534  & 4,466,702 & 9,457,678 & 1,083,140 & 4,725,764 & 10,500,100  \\ \hline

    \end{tabular}
    }
\label{Tab: Parameters}
\end{table}

\subsection{Action Subsampling}
All modules presented use actions as features at some point within their architecture, and predict a reward for $k_f$ timesteps conditioned on these actions.
At each timestep, the agent is provided with a random subsample of the complete 224x224 = 50176 pixel action space such that $a \sim U(0,224)$.
For SR and MTS tasks, the agent receives 1000 unique action samples, and is increased to 5000 for localization.
Only these subsamples participate in reward prediction and action sampling at this timestep.
The sampled action is the only one to be stored for future time steps or appear in gradient updates.
Models learn by minimizing the sum of Cross Entropies of all $k_f$ predicted rewards.

\subsection{Action Choice Policy}
The output of the network at time $t$ are $k_f$ reward maps $m(x)$, contstructed by computing the expected reward obtained for each subsampled action, given current visual input and history $\mathbf{C}_t$ and action choice history $\mathbf{h}_{t-1}$ .

$$m(x) = \left\lbrace E\left[r_{t}|a_{j},\mathbf{h}_{t-1},\mathbf{C}_t\right] \right\rbrace_{j=1}^{j=N_{samples}} $$

Where each expected value is calculated as
$$E\left[r_{t}|a_{j},\mathbf{h}_{t-1},\mathbf{C}_t\right] = \sum_{i}r_{i}P(r_{t}=r_{i}|a_{j},\mathbf{h}_{t-1},\mathbf{C}_t)$$

For the present study,  SR and MTS tasks have $r_i \in \left\lbrace 0.0,1.0\right\rbrace$, and localization uses \linebreak$r_i \in \left\lbrace0.0, 0.5, 0.6, 0.7, 0.8, 0.9, 1.0\right\rbrace$

The action itself is chosen by sampling the map with the largest variance. 
$$a_t \sim \mathbf{VarArgmax}_{j=1}^{k_f} \{\mathbf{Dist}[\mathbf{Norm}[m_j]]\},$$ 
where $$Norm[m] = m - \min_{x \in \mathcal{A}} m(x)$$ removes the minimum of the map, and
$$Dist[m] = \frac{m}{\sum_{x \in \mathcal{A}} m(x)}$$ 
ensures it is a probability distribution and $\mathbf{VarArgmax}$ is an operator which chooses the input with largest variance. 

In experimenting with different exploration policies, we found that this sampling procedure was empirically superior.
Random $\epsilon$-greedy for instance is inefficient for large-action spaces, and becomes more efficient at the end of learning rather than at the beginning (an important distinction when considering task-switching).
An alternate version of $\epsilon$-greedy was devised in which it defaulted to our sampling policy rather than random choice, but performance still lacked. 
Other sampling policies were also attempted -- such as paramaterizing the map as a Boltzmann distribution rather than weighted uniform --  although this resulted in poor sampling since a larger proportion of probability mass is devoted to poor action choices.

\subsection{Hyperparameters}
Learning rates for the ADAM optimizer were chosen on a per-task basis through cross-validation on a grid between [$10^{-4}$,$10^{-3}$] for each architecture.
Values used in the present study may be seen in Table \ref{Tab: Learning-Rates}.

\begin{table}
 \centering
  \caption{Module learning rates}
  \tiny{
  \hspace*{-1.0cm}
    \begin{tabular}{l *{12}{p{0.8cm}}}
    
        ~           & 2-way SR  & 4-way swap SR & 4-way stationary SR & 2-way stationary MTS & 2-way vert-motion MTS & 2-way horiz flip MTS & 2-way motion/flip MTS & 4-way 2-shown MTS & 4-way 2-shown vert-motion MTS & 4-way 4-shown stationary MTS & 4-way 4-shown permuted MTS & LOC \\ \hline
        CReZ-CReS   & $10^{-3}$ & $10^{-3}$     & $10^{-3}$           & $5\cdot 10^{-4}$     & $5\cdot 10^{-4}$      & $5\cdot 10^{-4}$     & $5\cdot 10^{-4}$      & $5\cdot 10^{-4}$  & $5\cdot 10^{-4}$              & $5\cdot 10^{-4}$             & $5\cdot 10^{-4}$           & $10^{-4}$    \\ 
        CReZ-LRS    & $10^{-3}$ & $10^{-3}$     & $10^{-3}$           & $5\cdot 10^{-4}$     & $5\cdot 10^{-4}$      & $5\cdot 10^{-4}$     & $5\cdot 10^{-4}$      & $5\cdot 10^{-4}$  & $5\cdot 10^{-4}$              & $5\cdot 10^{-4}$             & $5\cdot 10^{-4}$           & $10^{-4}$    \\ 
        LRS         & $10^{-3}$ & $10^{-3}$     & $10^{-3}$           & $10^{-3}$            & $10^{-3}$             & $10^{-3}$            & $10^{-3}$             & $10^{-3}$         & $10^{-3}$                     & $10^{-3}$                    & $2\cdot10^{-4}$            & $10^{-4}$    \\ 
        LS          & $10^{-3}$ & $10^{-3}$     & $10^{-3}$           & $10^{-3}$            & $10^{-3}$             & $10^{-3}$            & $10^{-3}$             & $10^{-3}$         & $10^{-3}$                     & $10^{-3}$                    & $2\cdot10^{-4}$            & $10^{-4}$    \\ 
        LR          & $10^{-3}$ & $10^{-3}$     & $10^{-3}$           & $10^{-3}$            & $10^{-3}$             & $10^{-3}$            & $10^{-3}$             & $10^{-3}$         & $10^{-3}$                     & $10^{-3}$                    & $10^{-3}$                  & $10^{-4}$    \\ 
        LT          & $10^{-3}$ & $10^{-3}$     & $10^{-3}$           & $10^{-3}$            & $10^{-4}$             & $10^{-3}$            & $10^{-3}$             & $10^{-3}$         & $10^{-3}$                     & $10^{-4}$                    & $10^{-4}$                  & $10^{-4}$    \\ 
        LSig        & $10^{-3}$ & $10^{-3}$     & $10^{-3}$           & $10^{-3}$            & $10^{-4}$             & $10^{-3}$            & $10^{-3}$             & $10^{-3}$         & $10^{-3}$                     & $10^{-4}$                    & $10^{-4}$                  & $10^{-4}$    \\ 
        LE          & $10^{-3}$ & $10^{-3}$     & $10^{-4}$           & $10^{-3}$            & $10^{-3}$             & $10^{-3}$            & $10^{-3}$             & $10^{-3}$         & $10^{-3}$                     & $10^{-3}$                    & $10^{-3}$                  & $10^{-4}$    \\ 
        LCre        & $10^{-3}$ & $10^{-3}$     & $10^{-3}$           & $10^{-3}$            & $10^{-3}$             & $10^{-3}$            & $10^{-3}$             & $10^{-3}$         & $10^{-3}$                     & $10^{-3}$                    & $10^{-3}$                  & $10^{-4}$    \\ 
        LBR-small   & $10^{-3}$ & $10^{-3}$     & $10^{-3}$           & $10^{-3}$            & $10^{-3}$             & $10^{-3}$            & $10^{-3}$             & $10^{-3}$         & $10^{-3}$                     & $10^{-3}$                    & $10^{-3}$                  & $10^{-4}$    \\ 
        LBR-med     & $10^{-3}$ & $10^{-3}$     & $10^{-3}$           & $10^{-4}$            & $10^{-4}$             & $10^{-4}$            & $10^{-4}$             & $10^{-4}$         & $10^{-4}$                     & $10^{-4}$                    & $10^{-4}$                  & $10^{-4}$    \\ 
        LBR-large   & $10^{-3}$ & $10^{-3}$     & $10^{-4}$           & $10^{-4}$            & $10^{-4}$             & $10^{-4}$            & $10^{-4}$             & $10^{-4}$         & $10^{-4}$                     & $10^{-4}$                    & $10^{-4}$                  & $10^{-4}$    \\ 
        LBT-small   & $10^{-4}$ & $10^{-4}$     & $10^{-4}$           & $10^{-3}$            & $10^{-3}$             & $10^{-3}$            & $10^{-3}$             & $10^{-3}$         & $10^{-3}$                     & $10^{-4}$                    & $10^{-4}$                  & $10^{-4}$    \\ 
        LBT-med     & $10^{-4}$ & $10^{-4}$     & $10^{-4}$           & $10^{-3}$            & $10^{-3}$             & $10^{-3}$            & $10^{-3}$             & $10^{-3}$         & $10^{-3}$                     & $10^{-4}$                    & $10^{-4}$                  & $10^{-4}$    \\ 
        LBT-large   & $10^{-4}$ & $10^{-4}$     & $10^{-4}$           & $10^{-4}$            & $10^{-4}$             & $10^{-4}$            & $10^{-4}$             & $10^{-4}$         & $10^{-4}$                     & $10^{-4}$                    & $10^{-4}$                  & $10^{-4}$    \\ 
        LBSig-small & $10^{-4}$ & $10^{-4}$     & $10^{-4}$           & $10^{-3}$            & $10^{-3}$             & $10^{-3}$            & $10^{-3}$             & $10^{-3}$         & $10^{-3}$                     & $10^{-4}$                    & $10^{-3}$                  & $10^{-4}$    \\ 
        LBSig-med   & $10^{-4}$ & $10^{-4}$     & $10^{-4}$           & $10^{-3}$            & $10^{-3}$             & $10^{-3}$            & $10^{-3}$             & $10^{-3}$         & $10^{-3}$                     & $10^{-4}$                    & $10^{-4}$                  & $10^{-4}$    \\ 
        LBSig-large & $10^{-4}$ & $10^{-4}$     & $10^{-4}$           & $10^{-4}$            & $10^{-4}$             & $10^{-4}$            & $10^{-4}$             & $10^{-4}$         & $10^{-4}$                     & $10^{-4}$                    & $10^{-4}$                  & $10^{-4}$    \\ 
        LBE-small   & $10^{-3}$ & $10^{-3}$     & $10^{-3}$           & $10^{-3}$            & $10^{-3}$             & $10^{-3}$            & $10^{-3}$             & $10^{-3}$         & $10^{-3}$                     & $10^{-3}$                    & $10^{-3}$                  & $10^{-4}$    \\ 
        LBE-med     & $10^{-3}$ & $10^{-3}$     & $10^{-3}$           & $10^{-4}$            & $10^{-4}$             & $10^{-4}$            & $10^{-4}$             & $10^{-4}$         & $10^{-4}$                     & $10^{-3}$                    & $10^{-4}$                  & $10^{-4}$    \\ 
        LBE-large   & $10^{-4}$ & $10^{-4}$     & $10^{-3}$           & $10^{-4}$            & $10^{-4}$             & $10^{-4}$            & $10^{-4}$             & $10^{-4}$         & $10^{-4}$                     & $10^{-4}$                    & $10^{-4}$                  & $10^{-4}$    \\ 
        LBCre-small & $10^{-3}$ & $10^{-3}$     & $10^{-3}$           & $10^{-3}$            & $10^{-3}$             & $10^{-4}$            & $10^{-3}$             & $10^{-3}$         & $10^{-3}$                     & $10^{-3}$                    & $10^{-3}$                  & $10^{-4}$    \\ 
        LBCre-med   & $10^{-3}$ & $10^{-3}$     & $10^{-3}$           & $10^{-4}$            & $10^{-4}$             & $10^{-4}$            & $10^{-4}$             & $10^{-4}$         & $10^{-4}$                     & $10^{-3}$                    & $10^{-4}$                  & $10^{-4}$    \\ 
        LBCre-large & $10^{-4}$ & $10^{-4}$     & $10^{-4}$           & $10^{-4}$            & $10^{-4}$             & $10^{-4}$            & $10^{-4}$             & $10^{-4}$         & $10^{-4}$                     & $10^{-4}$                    & $10^{-4}$                  & $10^{-4}$    \\
   \\ \hline

    \end{tabular}
    }
\label{Tab: Learning-Rates}
\end{table}

\section{Additional Learning Curves}

\begin{figure}
\centering
\includegraphics [width=1\linewidth]{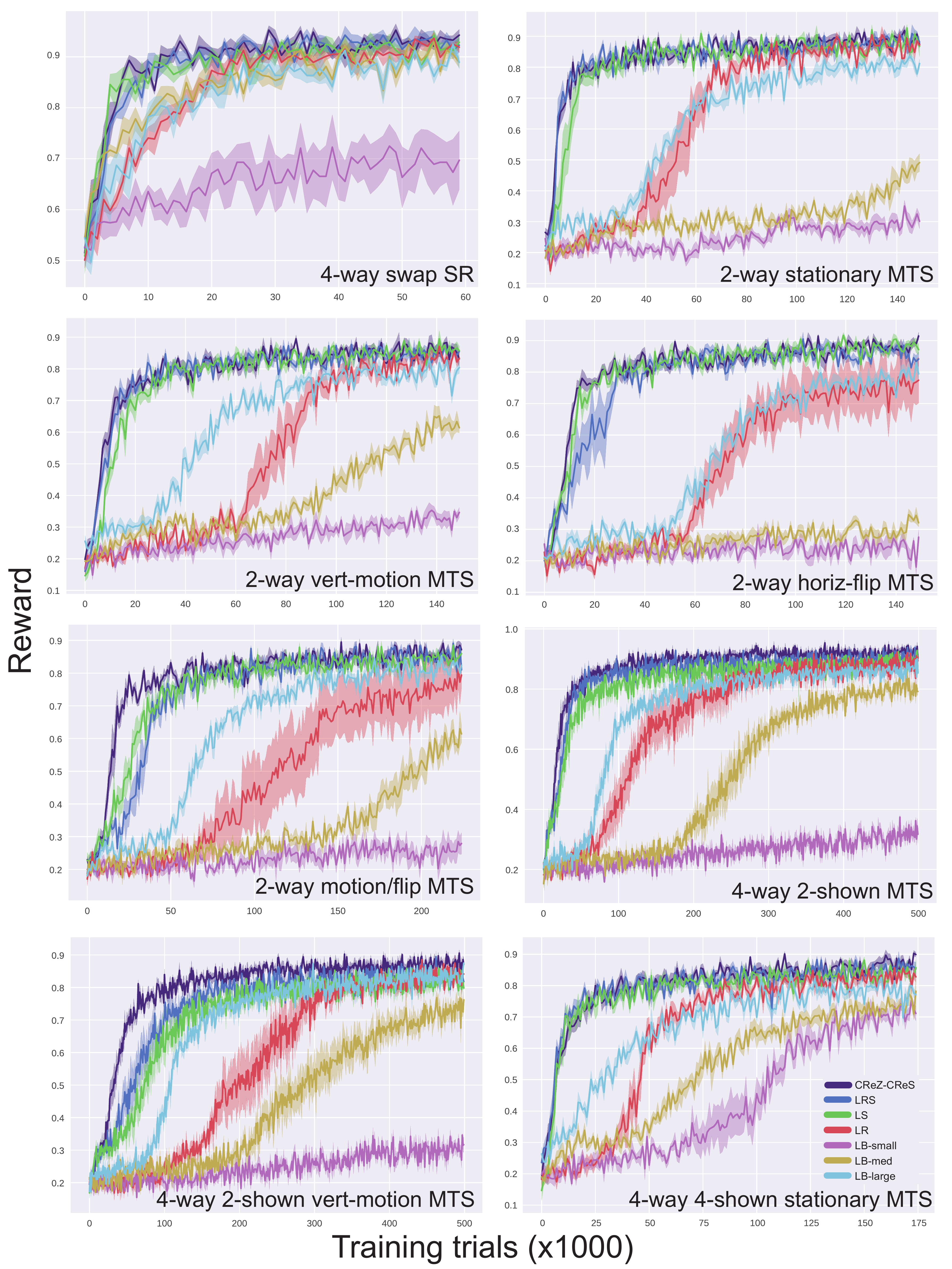}
\vspace{-2mm}
\caption{\textbf{Learning curves.} for \textbf{a.} four-way double binary SR, \textbf{b.} two-way stationary MTS, \textbf{c.} two-way MTS with random vertical motion, \textbf{d.} two-way MTS with random horizontal match image flips, \textbf{e.} two-way MTS with random vertical motion and  horizontal flips, \textbf{f.} four-way two-shown MTS, \textbf{g.} four-way two-shown MTS with vertical motion, and \textbf{h.} four-way four-shown MTS with stationary ordering.}
\label{fig:curves}
\end{figure}

Learning trajectories for eight additional tasks are provided in Figure \ref{fig:curves}.
Modules capable of convergence on a task were run until this was achieved, but AUC values for a given task are calculated at the point in time when the majority of models converge.

\end{document}